\pdfoutput=1

\documentclass[11pt]{article}

\usepackage{emnlp2021}

\usepackage{times}
\usepackage{latexsym}
\usepackage{times}
\usepackage{arydshln}
\usepackage{booktabs}
\usepackage{float}
\usepackage{latexsym}
\usepackage{tabularx}
\usepackage[ruled, vlined, linesnumbered]{algorithm2e}
\usepackage{setspace}
\usepackage{subfigure}   
\usepackage{multirow}
\usepackage{graphicx}
\usepackage{amsfonts}
\usepackage{amsmath}
\usepackage[T1]{fontenc}

\usepackage[utf8]{inputenc}

\usepackage{microtype}

\title{Smoothing Dialogue States for Open Conversational Machine Reading}

\author{Zhuosheng Zhang\textsuperscript{1,2,3,\#}, Siru Ouyang\textsuperscript{1,2,3,\#}, Hai Zhao\textsuperscript{1,2,3,\thanks{\ \ Corresponding author. \# Equal contribution. This paper was partially supported by Key Projects of National Natural Science Foundation of China (U1836222 and 61733011).}},
Masao Utiyama\textsuperscript{4}, Eiichiro Sumita\textsuperscript{4}\\
\textsuperscript{1} Department of Computer Science and Engineering, Shanghai Jiao Tong University\\
\textsuperscript{2} Key Laboratory of Shanghai Education Commission for Intelligent Interaction\\
and Cognitive Engineering, Shanghai Jiao Tong University\\
\textsuperscript{3}MoE Key Lab of Artificial Intelligence, AI Institute, Shanghai Jiao Tong University\\
\textsuperscript{4}National Institute of Information and Communications Technology (NICT), Kyoto, Japan \\
\texttt{\{zhangzs,oysr0926\}@sjtu.edu.cn,zhaohai@cs.sjtu.edu.cn}\\
\texttt{\{mutiyama,eiichiro.sumita\}@nict.go.jp}\\
}

\begin{document}
\maketitle
\begin{abstract}
Conversational machine reading (CMR) requires machines to communicate with humans through multi-turn interactions between two salient dialogue states of decision making and question generation processes. In open CMR settings, as the more realistic scenario, the retrieved background knowledge would be noisy, which results in severe challenges in the information transmission. Existing studies commonly train independent or pipeline systems for the two subtasks. However, those methods are trivial by using hard-label decisions to activate question generation, which eventually hinders the model performance. In this work, we propose an effective gating strategy by smoothing the two dialogue states in only one decoder and bridge decision making and question generation to provide a richer dialogue state reference. Experiments on the OR-ShARC dataset show the effectiveness of our method, which achieves new state-of-the-art results. 



\end{abstract}

\section{Introduction}
The ultimate goal of multi-turn dialogue is to enable the machine to interact with human beings and solve practical problems \cite{zhu2018lingke,zhang2018modeling,zaib2020short,huang2020challenges,fan2020survey,gu-etal-2021-mpc}. It usually adopts the form of question answering (QA) according to the user's query along with the dialogue context \cite{sun2019dream,reddy2019coqa,choi2018quac}. The machine may also actively ask questions for confirmation \cite{Wu2017Neural,cai-etal-2019-skeleton,DialoGPTLG2020Zhang,gu2020dialogbert}. 

In the classic spoken language understanding tasks \cite{tur2011spoken,zhang2020graph,ren2018towards,qin2021survey}, specific slots and intentions are usually defined. According to these predefined patterns, the machine interacts with people according to the dialogue states, and completes specific tasks, such as ordering meals \cite{liu2013asgard} and air tickets \cite{price1990evaluation}. In real-world scenario, annotating data such as intents and slots is expensive. Inspired by the studies of reading comprehension \cite{rajpurkar2016squad,rajpurkar2018know,zhang2020semantics,zhang2021retrospective}, there appears a more general task --- conversational machine reading (CMR) \cite{saeidi-etal-2018-interpretation}: given the inquiry, the machine is required to retrieve relevant supporting rule documents, the machine should judge whether the goal is satisfied according to the dialogue context, and make decisions or ask clarification questions. 

A variety of methods have been proposed for the CMR task, including 1) sequential models that encode all the elements and model the matching relationships with attention mechanisms \cite{zhong-zettlemoyer-2019-e3,lawrence-etal-2019-attending,verma-etal-2020-neural,gao-etal-2020-explicit,gao-etal-2020-discern}; 2) graph-based methods that capture the discourse structures of the rule texts and user scenario for better interactions \cite{ouyang2020dialogue}. However, there are two sides of challenges that have been neglected:

\begin{figure*}[!t]
\centering
\includegraphics[width=1\textwidth]{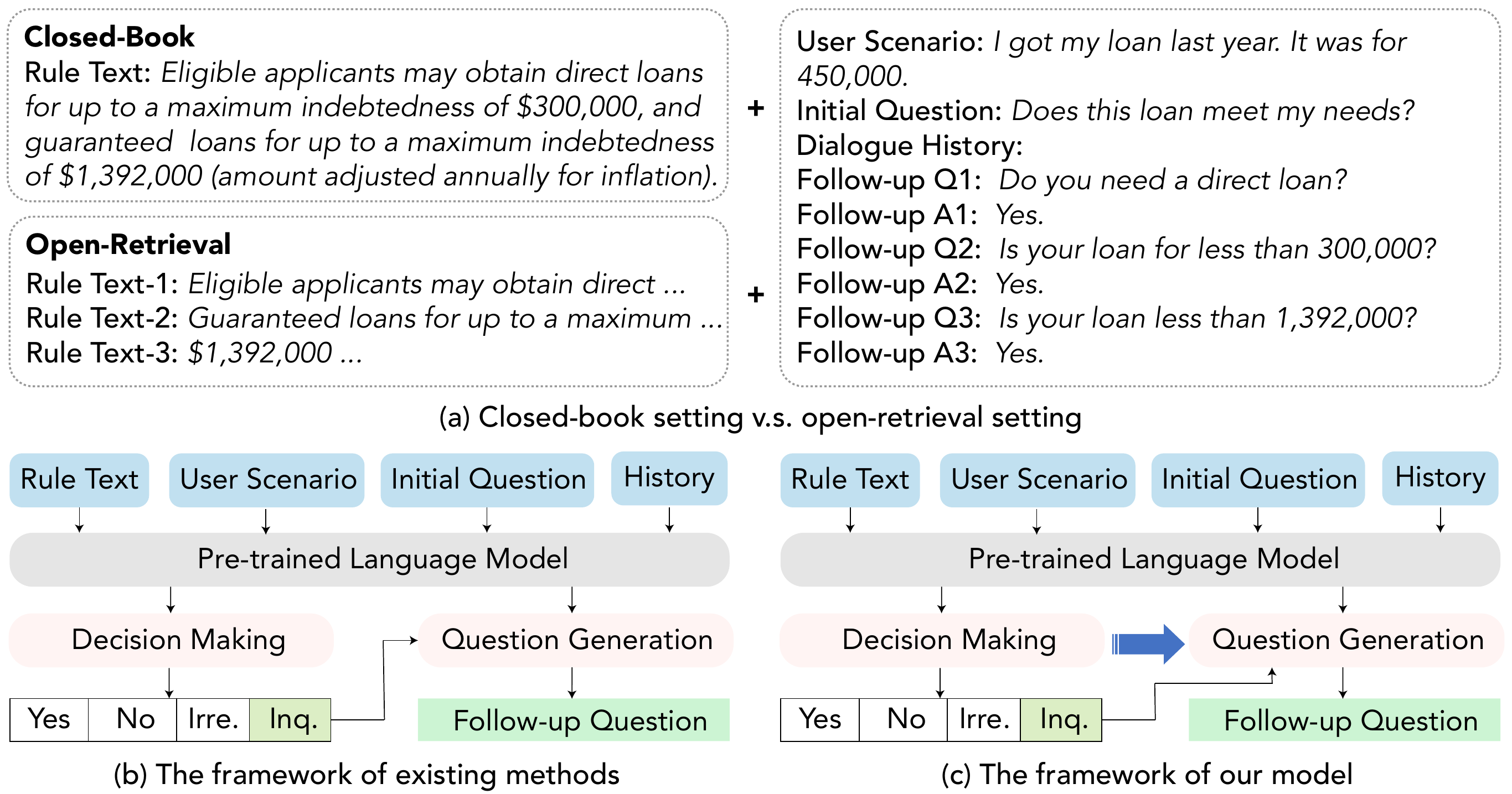}
\caption{The overall framework for our proposed model (c) compared with the existing ones (b). Previous studies generally regard CMR as two separate tasks and design independent systems. Technically, only the result of decision making will be fed to the question generation module, thus there is a gap between the dialogue states of decision making and question generation. To reduce the information gap, our model bridges the information transition between the two salient dialogue states and benefits from a richer rule reference through open-retrieval (a).}
\label{model-framework}
\end{figure*}

1) Open-retrieval of supporting evidence. The above existing methods assume that the relevant rule documents are given before the system interacts with users, which is in a closed-book style. In real-world applications, the machines are often required to retrieve supporting information to respond to incoming high-level queries in an interactive manner, which results in an open-retrieval setting \cite{gao2021open}. The comparison of the closed-book setting and open-retrieval setting is shown in Figure \ref{model-framework}.

2) The gap between decision making and question generation. Existing CMR studies generally regard CMR as two separate tasks and design independent systems. Only the result of decision making will be fed back to the question generation module. As a result, the question generation module knows nothing about the actual conversation states, which leads to poorly generated questions. There are even cases when the decision masking result is improved, but the question generation is decreased as reported in previous studies \cite{ouyang2020dialogue}. 

In this work, we design an end-to-end system by \underline{O}pen-retrieval of \underline{S}upporting evidence and bridging de\underline{C}ision m\underline{A}king and question gene\underline{R}ation (\textsc{Oscar}),\footnote{Our source codes are available at \url{https://github.com/ozyyshr/OSCAR}.} to bridge the information transition  between the two salient dialogue states of decision making and question generation, at the same time benefiting from a richer rule reference through open retrieval. In summary, our contributions are three folds:

1) For the task, we investigate the open-retrieval setting for CMR. We bridge decision making and question generation for the challenging CMR task, which is the first practice to our best knowledge.

2) For the technique, we design an end-to-end framework where the dialogue states for decision making are employed for question generation, in contrast to the independent models or pipeline systems in previous studies. Besides, a variety of strategies are empirically studied for smoothing the two dialogue states in only one decoder.

3) Experiments on the ShARC dataset show the effectiveness of our model, which achieves the new state-of-the-art results. A series of analyses show the contributing factors.

\begin{figure*}[!t]\label{model_overview}
\centering
\subfigure{
\begin{minipage}[t]{0.35\linewidth}
\centering
\includegraphics[width=1.0\textwidth]{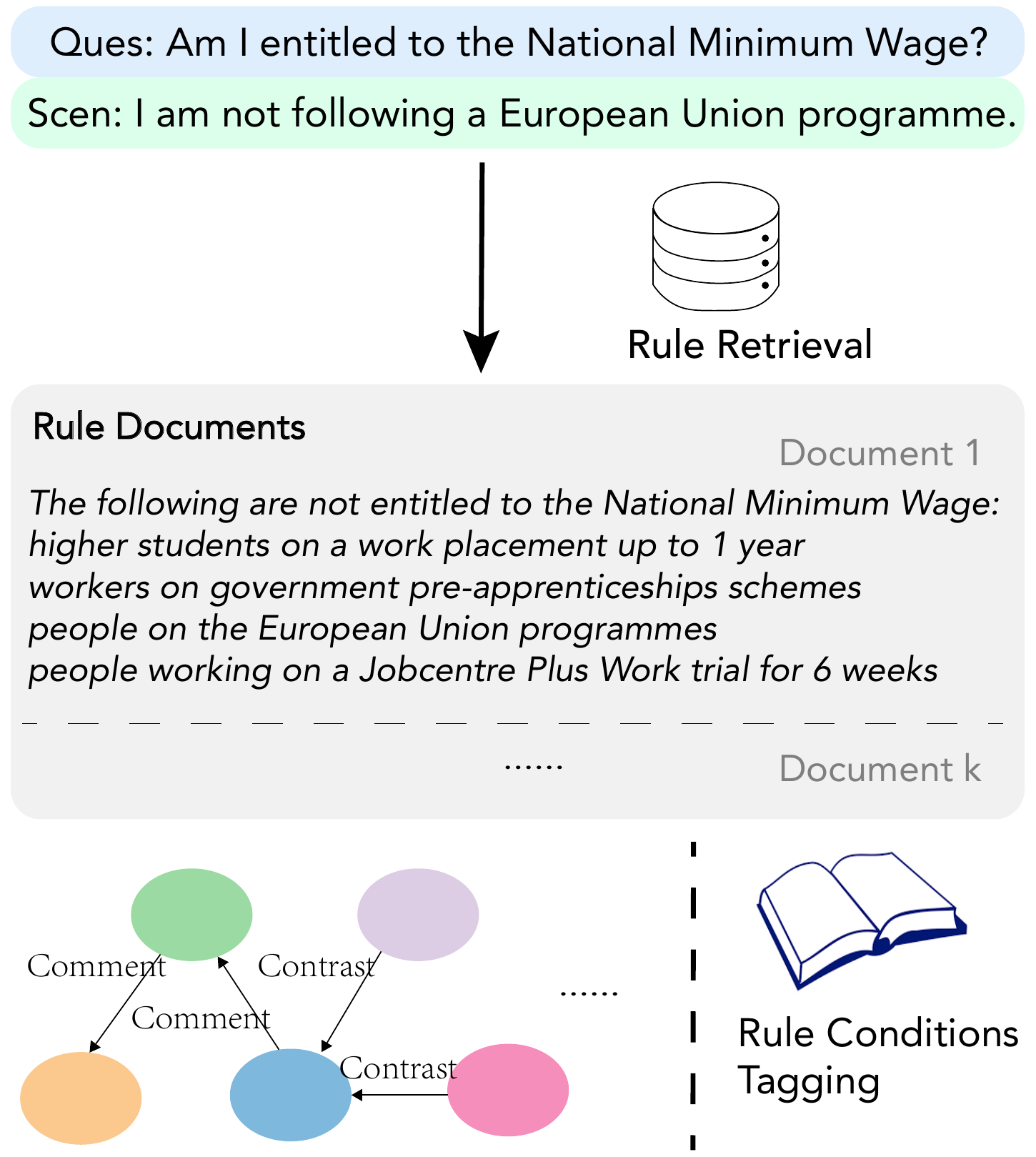}
\end{minipage}%
}%
\subfigure{
\begin{minipage}[t]{0.65\linewidth}
\centering
\includegraphics[width=1.0\textwidth]{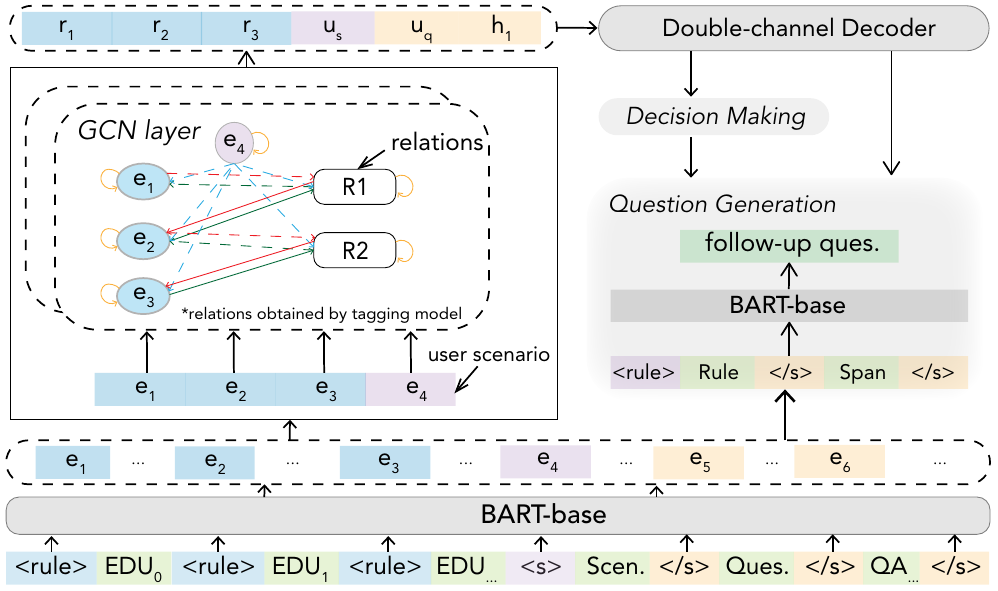}
\end{minipage}
}
\caption{The overall structure of our model \textsc{Oscar}. The left part introduces the retrieval and tagging process for rule documents, which is then fed into the encoder together with other necessary information.}
\label{model-overview}
\end{figure*}

\section{Related Work}
Most of the current conversation-based reading comprehension tasks are formed as either span-based QA \cite{reddy2019coqa,choi2018quac} or multi-choice tasks \cite{sun2019dream,mutual}, both of which neglect the vital process of question generation for confirmation during the human-machine interaction. In this work, we are interested in building a machine that can not only make the right decisions but also raise questions when necessary. The related task is called conversational machine reading \citep{saeidi-etal-2018-interpretation} which consists of two separate subtasks: decision making and question generation. Compared with conversation-based reading comprehension tasks, our concerned CMR task is more challenging as it involves rule documents, scenarios, asking clarification questions, and making a final decision. 

Existing works \cite{zhong-zettlemoyer-2019-e3,lawrence-etal-2019-attending,verma-etal-2020-neural,gao-etal-2020-explicit,gao-etal-2020-discern,ouyang2020dialogue} have made progress in modeling the matching relationships between the rule document and other elements such as user scenarios and questions. 
These studies are based on the hypothesis that 
the supporting information for answering the question is provided, which does not meet the real-world applications. Therefore, we are motivated to investigate the open-retrieval settings \cite{qu2020open}, where the retrieved background knowledge would be noisy. \citet{gao2021open} makes the initial attempts of open-retrieval for CMR. However, like previous studies, 
the common solution is training independent or pipeline systems for the two subtasks and does not consider the information flow between decision making and question generation, which would eventually hinder the model performance. Compared to existing methods, our method makes the first attempt to bridge the gap between decision making and question generation, by smoothing the two dialogue states in only one decoder. In addition, we improve the retrieval process by taking advantage of the traditional TF-IDF method and the latest dense passage retrieval model \cite{karpukhin2020dense}.





\section{Open-retrieval Setting for CMR}
In the CMR task, each example is formed as a tuple $\{R, U_s, U_q, C\}$, where $R$ denotes the rule texts, $U_s$ and $U_q$ are user scenarios and user questions, respectively, and $C$ represents the dialogue history. For open-retrieval CMR, $R$ is a subset retrieved from a large candidate corpus $\mathcal{D}$. The goal is to train a discriminator $ \mathcal{F(\cdot,\cdot)} $ for decision making, and a generator $ \mathcal{G(\cdot,\cdot)} $ on $\{R, U_s, U_q, C\}$ for question generation.

\section{Model}
Our model is composed of three main modules: retriever, encoder, and decoder. The retriever is employed to retrieve the related rule texts for the given user scenario and question. The encoder takes the tuple $\{R, U_s, U_q, C\}$ as the input, encodes the elements into vectors and captures the contextualized representations. The decoder makes a decision or generates a question once the decision is ``inquiry''. Figure \ref{model-framework} overviews the model architecture, we will elaborate the details in the following part. 

\subsection{Retrieval}
To obtain the supporting rules, we construct the query by concatenating the user question and user scenario. The retriever calculates the semantic matching score between the query and the candidate rule texts from the pre-defined corpus and returns the top-$k$ candidates. In this work, we employ TF-IDF and DPR \cite{karpukhin2020dense} in our retrieval, which are representatives for sparse and dense retrieval methods. TF-IDF stands for term frequency-inverse document frequency, which is used to reflect how relevant a term is in a given document. DPR is a dense passage retrieval model that calculates the semantic matching using dense vectors, and it uses embedding functions that can be trained for specific tasks.

\subsection{Graph Encoder}\label{model:encoding}
One of the major challenges of CMR is interpreting rule texts, which have complex logical structures between various inner rule conditions. According to Rhetorical Structure Theory (RST)  of discourse parsing \cite{mann1988rhetorical}, we utilize a pre-trained discourse parser \cite{shi2019deep}\footnote{This discourse parser gives a state-of-the-art performance on STAC so far. There are $16$ discourse relations according to STAC \citep{asher-etal-2016-discourse}, including comment, clarification-question, elaboration, acknowledgment, continuation, explanation, conditional, question-answer, alternation, question-elaboration, result, background, narration, correction, parallel, and contrast.} to break the rule text into clause-like units called elementary discourse units (EDUs) to extract the in-line rule conditions from the rule texts.  


\paragraph{Embedding}
    We employ pre-trained language model (PrLM) model as the backbone of the encoder. As shown in the figure, the input of our model includes rule document which has already be parsed into EDUs with explicit discourse relation tagging, user initial question, user scenario and the dialog history. Instead of inserting a \texttt{[CLS]} token before each rule condition to get a sentence-level representation, we use \texttt{[RULE]} which is proved to enhance performance \citep{lee-etal-2020-slm}. Formally, the sequence is organized as: \{\texttt{[RULE]} EDU$_0$ \texttt{[RULE]} EDU$_1$ \texttt{[RULE]} EDU$_k$ \texttt{[CLS]} Question \texttt{[CLS]} Scenario \texttt{[CLS]} History \texttt{[SEP]}\}.
    Then we feed the sequence to the PrLM to obtain the contextualized representation.

\paragraph{Interaction}
To explicitly model the discourse structure among the rule conditions, we first annotate the discourse relationships between the rule conditions and employ a relational graph convolutional network following \citet{ouyang2020dialogue} by regarding the rule conditions as the vertices. The graph is formed as a Levi graph \citep{levi-1942} that regards the relation edges as additional vertices. For each two vertices, there are six types of possible edges derived from the discourse parsing, namely, \textit{default-in}, \textit{default-out}, \textit{reverse-in}, \textit{reverse-out}, \textit{self}, and \textit{global}. Furthermore, to build the relationship with the background user scenario, we add an extra global vertex of the user scenario that connects all the other vertices. As a result, there are three types of vertices, including the rule conditions, discourse relations, and the global scenario vertex. 

For rule condition and user scenario vertices, we fetch the contextualized representation of the special tokens \texttt{[RULE]} and \texttt{[CLS]} before the corresponding sequences, respectively. For relation vertices, they are initialized as the conventional embedding layer, whose representations are obtained through a lookup table.

For each rule document that is composed of multiple rule conditions, i.e., EDUs, let $h_p$ denote the initial representation of every node $v_p$, the graph-based information flow process can be written as:
\begin{equation}
    h_p^{(l+1)}=\textup{ReLU}(\sum_{r\in R_L}\sum_{v_q\in \mathcal{N}_r(v_p)}\frac{1}{c_{p,r}}w_r^{(l)}h_q^{(l)}),
\end{equation}
where $\mathcal{N}_r(v_p)$ denotes the neighbors of node $v_p$ under relation $r$ and $c_{p,r}$ is the number of those nodes. $w_r^{(l)}$ is the trainable parameters of layer $l$.

We have the last-layer output of discourse graph:
\begin{equation}
\begin{split}
    g_{p}^{(l)}&=\textup{Sigmoid}(h_p^{(l)}W_{r,g}), \\
    r_p^{(l+1)}&=\textup{ReLU}(\sum_{r\in R_L}\sum_{v_q\in \mathcal{N}_r(v_p)}g_{q}^{(l)}\frac{1}{c_{p,r}}w_r^{(l)}h_q^{(l)}),
    \end{split}    
\end{equation}
where $W^{(l)}_{r,g}$ is a learnable parameter under relation type $r$ of the $l$-th layer. The last-layer hidden states for all the vertices $r_p^{(l+1)}$ are used as the graph representation for the rule document. For all the $k$ rule documents from the retriever, we concatenate $r_p^{(l+1)}$ for each rule document, and finally have $r = \{r_1, r_2, \dots, r_m \}$ where $m$ is the total number of the vertices among those rule documents.

\subsection{Double-channel Decoder}\label{decoder}
Before decoding, we first accumulate all the available information through a self-attention layer \citep{NIPS2017_3f5ee243} by allowing all the rule conditions and other elements to attend to each other. Let $[r_1,r_2,\dots,r_m;u_q;u_s;h_1,h_2,\dots,h_n]$ denote all the representations, $r_i$ is the representation of the discourse graph, $u_q$, $u_s$ and $h_i$ stand for the representation of user question, user scenario and dialog history respectively. $n$ is the number of history QAs. After encoding, the output is represented as:
\begin{equation}
    H_c = [\tilde{r}_1,\tilde{r}_2,\dots,\tilde{r}_m;\tilde{u}_q,\tilde{u}_s;\tilde{h}_1,\tilde{h}_2,\dots, \tilde{h}_n],
\end{equation}
which is then used for the decoder.

\paragraph{Decision Making}Similar to existing works \citep{zhong-zettlemoyer-2019-e3, gao-etal-2020-explicit, gao-etal-2020-discern}, we apply an entailment-driven approach for decision making. A linear transformation tracks the fulfillment state of each rule condition among \textit{entailment}, \textit{contradiction} and \textit{Unmentioned}. As a result, our model makes the decision by
\begin{equation}
    f_i=W_f\tilde{r_i}+b_f\in \mathbb{R}^{3},
\end{equation}
where $f_i$ is the score predicted for the three labels of the $i$-th condition. This prediction is trained via a cross entropy loss for multi-classification problems:
\begin{equation}
    \mathcal{L}_{entail}=-\frac{1}{N}\sum_{i=1}^{N}\log \textup{softmax}(f_i)_r,
\end{equation}
where $r$ is the ground-truth state of fulfillment.

After obtaining the state of every rule, we are able to give a final decision towards whether it is \textsl{Yes, No, Inquire} or \textsl{Irrelevant} by attention.
\begin{equation}
\begin{aligned}
    \alpha_i&=w_{\alpha}^T[f_i;\tilde{r_i}]+b_\alpha \in \mathbb{R}^1,\\
    \tilde{\alpha_i}&=\textup{softmax}(\alpha)_i\in [0,1],\\
    z&=W_z\sum_i\tilde{\alpha_i}[f_i;\tilde{r_i}]+b_z\in \mathbb{R}^4,
\end{aligned}
\label{dec}
\end{equation}
where $\alpha_i$ is the attention weight for the $i$-th decision and $z$ has the score for all the four possible states. The corresponding training loss is
\begin{equation}
    \mathcal{L}_{decision}=-\log \textup{softmax}(z)_l.
\end{equation}

The overall loss for decision making is:
\begin{equation}
\mathcal{L}_d =\mathcal{L}_{decision} + \lambda\mathcal{L}_{entail}.
\end{equation}

\paragraph{Question Generation} If the decision is made to be \textsl{Inquire}, the machine needs to ask a follow-up question to further clarify. Question generation in this part is mainly based on the uncovered information in the rule document, and then that information will be rephrased into a question. We predict the position of an under-specified span within a rule document in a supervised way. Following \citet{devlin-etal-2019-bert}, our model learns a start vector $w_s\in \mathbb{R}^d$ and end vector $w_e\in \mathbb{R}^d$ to indicate the start and end positions of the desired span:
\begin{equation}
    span=\mathop{\arg\min}_{i,j,k} (w_s^Tt_{k,i}+w_e^Tt_{k,j}),
\end{equation}
where $t_{k,i}$ denote the $i$-th token in the $k$-th rule sentence. The ground-truth span labels are generated by calculating the edit distance between the rule span and the follow-up questions. Intuitively, the shortest rule span with the minimum edit distance is selected to be the under-specified span. 

Existing studies deal with decision making and question generation independently \cite{zhong-zettlemoyer-2019-e3,lawrence-etal-2019-attending,verma-etal-2020-neural,gao-etal-2020-explicit,gao-etal-2020-discern}, and use hard-label decisions to activate question generation. These methods inevitably suffer from error propagation if the model makes the wrong decisions. For example, if the made decision is not ``inquiry", the question generation module will not be activated which may be supposed to ask questions in the cases. For the open-retrieval CMR that involves multiple rule texts, it even brings more diverse rule conditions as a reference, which would benefit for generating meaningful questions.

Therefore, we concatenate the rule document and the predicted span to form an input sequence: $x$ = \texttt{[CLS]} Span \texttt{[SEP]} Rule Documents \texttt{[SEP]}. We feed $x$ to BART encoder \citep{NEURIPS2019_c20bb2d9} and obtain the encoded representation $H_e$. To take advantage of the contextual states of the overall interaction of the dialogue states, we explore two alternative smoothing strategies:
\begin{enumerate}
    \item \textbf{Direct Concatenation} concatenates $H_c$ and $H_e$ to have $H = [H_c; H_e]$.
    \item  \textbf{Gated Attention} applies multi-head attention mechanism \cite{NIPS2017_7181} to append the contextual states to $H_e$ to get $\hat H=Attn(H_e, K, V)$ where \{K,V\} are packed from $H_c$. Then a gate control $\lambda$ is computed as $sigmoid(W_{\lambda}\hat H+U_{\lambda}H^e)$ to get the final representation $H=H^e+\lambda \hat H$.
\end{enumerate}

$H$ is then passed to the BART decoder to generate the follow-up question. At the $i$-th time-step, $H$ is used to generate the target token $y_i$ by
\begin{equation}
    P(y_i\mid y_{<i}, x; \theta) \propto  \exp(W_d\tanh(W_w H)),
\end{equation}
where $\theta$ denotes all the trainable parameters. $W_d$ and $W_w$ are projection matrices. The training objective is computed by
\begin{equation}
    \mathcal{L}_g = \arg\max \sum^{I}_{i=1} \log P(y_i\mid y_{<i}, x;\theta).
\end{equation}

The overall loss function for end-to-end training is
\begin{equation}
    \mathcal{L} = \mathcal{L}_d + \mathcal{L}_g.
\end{equation}

\begin{table*}
\small
\centering\centering\setlength{\tabcolsep}{8.0pt}
\begin{tabular}{lcccccccc}
\toprule
\multirow{3}{*}{Model} &
\multicolumn{4}{c}{Dev Set} & \multicolumn{4}{c}{Test Set}\\
&\multicolumn{2}{c}{Decision Making} & \multicolumn{2}{c}{Question Gen.} & \multicolumn{2}{c}{Decision Making} & \multicolumn{2}{c}{Question Gen.}\\
\cmidrule{2-5}
\cmidrule{6-9}
 & Micro & Macro & $\text{F1}_{\text{BLEU1}}$ & $\text{F1}_{\text{BLEU4}}$ & Micro & Macro & $\text{F1}_{\text{BLEU1}}$ & $\text{F1}_{\text{BLEU4}}$ \\ 
\midrule
\textit{w/ TF-IDF} \\
 E$^3$ & 61.8\scriptsize{$\pm$0.9} & 62.3\scriptsize{$\pm$1.0} & 29.0\scriptsize{$\pm$1.2} & 18.1\scriptsize{$\pm$1.0}&  61.4\scriptsize{$\pm$2.2} & 61.7\scriptsize{$\pm$1.9} & 31.7\scriptsize{$\pm$0.8} & 22.2\scriptsize{$\pm$1.1}\\
 EMT & 65.6\scriptsize{$\pm$1.6} & 66.5\scriptsize{$\pm$1.5} & 36.8\scriptsize{$\pm$1.1}& 32.9\scriptsize{$\pm$1.1}&  64.3\scriptsize{$\pm$0.5} & 64.8\scriptsize{$\pm$0.4} & 38.5\scriptsize{$\pm$0.5} & 30.6\scriptsize{$\pm$0.4} \\
 DISCERN & 66.0\scriptsize{$\pm$1.6} & 66.7\scriptsize{$\pm$1.8} & 36.3\scriptsize{$\pm$1.9} & 28.4\scriptsize{$\pm$2.1} &  66.7\scriptsize{$\pm$1.1} & 67.1\scriptsize{$\pm$1.2} & 36.7\scriptsize{$\pm$1.4} & 28.6\scriptsize{$\pm$1.2} \\
 DP-RoBERTa & 73.0\scriptsize{$\pm$1.7} & 73.1\scriptsize{$\pm$1.6} & 45.9\scriptsize{$\pm$1.1} & 40.0\scriptsize{$\pm$0.9} &  70.4\scriptsize{$\pm$1.5} & 70.1\scriptsize{$\pm$1.4} & 40.1\scriptsize{$\pm$1.6} & 34.3\scriptsize{$\pm$1.5} \\
 MUDERN & 78.4\scriptsize{$\pm$0.5} & 78.8\scriptsize{$\pm$0.6} & 49.9\scriptsize{$\pm$0.8} & 42.7\scriptsize{$\pm$0.8}&  75.2\scriptsize{$\pm$1.0} & 75.3\scriptsize{$\pm$0.9} & 47.1\scriptsize{$\pm$1.7} & 40.4\scriptsize{$\pm$1.8}\\ 
\cdashline{1-9}
 \textit{w/ DPR++} \\
 MUDERN & 79.7\scriptsize{$\pm$1.2} & 80.1\scriptsize{$\pm$1.0} &50.2\scriptsize{$\pm$0.7} &42.6\scriptsize{$\pm$0.5}\scriptsize{} & 75.6\scriptsize{$\pm$0.4} & 75.8\scriptsize{$\pm$0.3} & 48.6\scriptsize{$\pm$1.3}&40.7\scriptsize{$\pm$1.1}\\
\textsc{Oscar}& 80.5\scriptsize{$\pm$0.5} & 80.9\scriptsize{$\pm$0.6}  & 51.3\scriptsize{$\pm$0.8} & 43.1\scriptsize{$\pm$0.8}& 76.5\scriptsize{$\pm$0.5} & 76.4\scriptsize{$\pm$0.4} & 49.1\scriptsize{$\pm$1.1} & 41.9\scriptsize{$\pm$1.8} \\

\bottomrule
\end{tabular}
\caption{Results on the validation and test set of OR-ShARC. The first block presents the results of public models from \citet{gao2021open}, and the second block reports the results of our implementation of the SOTA model MUDERN, and ours based on DPR++. The average
results with a standard deviation on 5 random seeds are reported.}\label{table:e2e}
\end{table*}

\begin{table}
\small
\setlength{\tabcolsep}{3.6pt}
\centering\centering
\begin{tabular}{lcccc}
\toprule
\multirow{2}{*}{Model} &
\multicolumn{2}{c}{Seen} & \multicolumn{2}{c}{Unseen}\\
 & $\text{F1}_{\text{BLEU1}}$ & $\text{F1}_{\text{BLEU4}}$  & $\text{F1}_{\text{BLEU1}}$ & $\text{F1}_{\text{BLEU4}}$ \\ \midrule
 \textsc{MUDERN} & 62.6  & 57.8 & 33.1 &24.3 \\
\textsc{Oscar} & 64.6  &  59.6 & 34.9 &25.1  \\
\bottomrule
\end{tabular}
\caption{The comparison of question generation on the seen and unseen splits.}\label{table: seen-unseen}
\end{table}

\section{Experiments}\label{sec:experiment}
\subsection{Datasets} \label{sec:data}
For the evaluation of open-retrieval setting, we adopt the OR-ShARC dataset \citep{gao2021open}, which is a revision of the current CMR benchmark --- ShARC \cite{saeidi-etal-2018-interpretation}. The original dataset contains up to 948 dialog trees clawed from government websites. Those dialog trees are then flattened into 32,436 examples consisting of \textit{utterance\_id}, \textit{tree\_id}, \textit{rule document}, \textit{initial question}, \textit{user scenario}, \textit{dialog history}, \textit{evidence} and the \textit{decision}. The update of OR-ShARC is the removal of the gold rule text for each sample. Instead, all rule texts used in the ShARC dataset are served as the supporting knowledge sources for retrieval. There are 651 rules in total. Since the test set of ShARC is not public, the train, dev and test are further manually split, whose sizes are 17,936, 1,105, 2,373, respectively. For the dev and test sets, around 50\% of the samples ask questions on rule texts used in training (seen) while the remaining of them contain questions on unseen (new) rule texts. The rationale behind seen and unseen splits for the validation and test set is that the two cases mimic the real usage scenario: users may ask questions about rule text which 1) exists in the training data (i.e., dialog history, scenario) as well as 2) completely newly added rule text.

\subsection{Evaluation Metrics} 
For the decision-making subtask, ShARC evaluates the Micro- and Macro- Acc. for the results of classification. For question generation, the main metric is $\text{F1}_{\text{BLEU}}$ proposed in \citet{gao2021open}, which calculates the BLEU scores for question generation when the predicted decision is ``inquire".





\begin{table*}
\small
\centering
\setlength{\tabcolsep}{11.0pt}
\begin{tabular}{lcccccccc}
\toprule
\multirow{2}{*}{Model} &
\multicolumn{4}{c}{Dev Set} & \multicolumn{4}{c}{Test Set}\\
 & Top1 & Top5 & Top10 & Top20 & Top1 & Top5 & Top10 & Top20 \\ 
 \midrule
 TF-IDF & 53.8 & 83.4 & 94.0  & 96.6 &  66.9& 90.3 & 94.0 & 96.6 \\
 DPR & 48.1 & 74.6 & 84.9 & 90.5 & 52.4 & 80.3 & 88.9 & 92.6 \\
 TF-IDF + DPR & 66.3  & 90.0 & 92.4 & 94.5 & 79.8 & 95.4& 97.1& 97.5\\
\bottomrule
\end{tabular}
\caption{Comparison of the open-retrieval methods. }\label{table:or-total}
\end{table*}

\subsection{Implementation Details}
Following the current state-of-the-art MUDERN model \cite{gao2021open} for open CMR, we employ BART \citep{NEURIPS2019_c20bb2d9} as our backbone model and the BART model serves as our baseline in the following sections. For open retrieval with DPR, we fine-tune DPR in our task following the same training process as the official implementation, with the same data format stated in the DPR GitHub repository.\footnote{\url{https://github.com/facebookresearch/DPR}} Since the data process requires hard negatives (\texttt{hard\_negative\_ctxs}), we constructed them using the most relevant rule documents (but not the gold) selected by TF-IDF and left the \texttt{negative\_ctxs} to be empty as it can be. For discourse parsing, we keep all the default parameters of the original discourse relation parser\footnote{\url{https://github.com/shizhouxing/DialogueDiscourseParsing}}, with F1 score achieving $55$.
The dimension of hidden states is $768$ for both the encoder and decoder. The training process uses Adam \citep{2014arXiv1412.6980K} for $5$ epochs with a learning rate set to $5$e-$5$. We also use gradient clipping with a maximum gradient norm of $2$, and a total batch size of $16$. The parameter $\lambda$ in the decision making objective is set to 3.0. For BART-based decoder for question generation, the beam size is set to $10$ for inference. We report the averaged result of five randomly run seeds with deviations.


\subsection{Results}

Table \ref{table:e2e} shows the results of \textsc{Oscar} and all the baseline models for the End-to-End task on the dev and test set with respect to the evaluation metrics mentioned above. Evaluating results indicate that \textsc{Oscar} outperforms the baselines in all of the metrics. In particular, it outperforms the public state-of-the-art model MUDERN by $1.3$\% in Micro Acc. and $1.1$\% in Macro Acc for the decision making stage on the test set. The question generation quality is greatly boosted via our approaches. Specifically, $\text{F1}_{\text{BLEU1}}$ and $\text{F1}_{\text{BLEU4}}$ are increased by $2.0\%$ and $1.5\%$ on the test set respectively. 

Since the dev set and test set have a 50\% split of user questions between seen and unseen rule documents as described in Section \ref{sec:data}, to analyze the performance of the proposed framework over seen and unseen rules, we have added a comparison of question generation on the seen and unseen splits as shown in Table \ref{table: seen-unseen}. The results show consistent gains for both of the seen and unseen splits. 

\begin{table}
\small
\centering
\setlength{\tabcolsep}{4.2pt}
\begin{tabular}{lcccc}
\toprule
TF-IDF & Top1 & Top5 & Top10 & Top20 \\ 
 \midrule
Train & 59.9 & 83.8 & 94.4  & 94.2 \\
Dev & 53.8 & 83.4 & 94.0 & 96.6  \\
\quad Seen Only  & 62.0  & 84.2 & 90.2 & 93.2 \\
\quad Unseen Only & 46.9 & 82.8 & 90.7 & 83.1 \\
Test & 66.9 & 90.3 & 94.0 & 96.6 \\
\quad Seen Only & 62.1 & 83.4 & 89.4 & 93.8 \\
\quad Unseen Only & 70.4 & 95.3 & 97.3 & 98.7 \\
\bottomrule
\end{tabular}
\caption{Retrieval Results of TF-IDF. }\label{table:or-tfidf}
\end{table}

\section{Analysis}
\subsection{Comparison of Open-Retrieval Methods}
We compare two typical retrievals methods, TF-IDF and Dense Passage Retrieval (DPR), which are widely-used traditional models from sparse vector space and recent dense-vector-based ones for open-domain retrieval, respectively. We also present the results of TF-IDF+DPR (denoted DPR++) following \citet{karpukhin2020dense}, using a linear combination of their scores as the new ranking function.

The overall results are present in Table \ref{table:or-total}. We see that TF-IDF performs better than DPR, and combining TF-IDF and DPR (DPR++) yields substantial improvements. To investigate the reasons, we collect the detailed results of the \textit{seen} and \textit{unseen} subsets for the dev and test sets, from which we observe that TF-IDF generally works  well on both the seen and unseen sets, while DPR is degraded on the unseen set. The most plausible reason would be that DPR is trained on the training set, it can only give better results on the \textit{seen} subsets because \textit{seen} subsets share the same rule texts for retrieval with the training set. However, DPR may easily suffer from over-fitting issues that result in the relatively weak scores on the \textit{unseen} sets. Based on the complementary merits, combining the two methods would take advantage of both sides, which achieves the best results finally.

\begin{table}
\small
\centering
\setlength{\tabcolsep}{4.2pt}
\begin{tabular}{lcccc}
\toprule
DPR & Top1 & Top5 & Top10 & Top20 \\ 
 \midrule
Train & 77.2 & 96.5 & 99.0  & 99.8 \\
Dev & 48.1 & 74.6 & 84.9 & 90.5  \\
\quad Seen Only  & 77.4  & 96.8 & 98.6 & 99.6 \\
\quad Unseen Only & 23.8 & 56.2 & 73.6 & 83.0 \\
Test & 52.4 & 80.3 & 88.9 & 92.6 \\
\quad Seen Only & 76.2 & 96.1 & 98.6 & 99.8 \\
\quad Unseen Only & 35.0 & 68.8 & 81.9 & 87.3 \\
\bottomrule
\end{tabular}
\caption{Retrieval Results of DPR. }\label{table:or-dpr}
\end{table}

\subsection{Decision Making}
By means of TF-IDF + DPR retrieval, we compare our model with the previous SOTA model MUDERN \cite{gao2021open} for comparison on the open-retrieval setting. According to the results in Table \ref{table:e2e}, we observe that our method can achieve a better performance than DISCERN, which indicates that the graph-like discourse modeling works well in the open-retrieval setting in general. 

\subsection{Question Generation}
\paragraph{Overall Results} We first compare the vanilla question generation with our method with encoder states. Table \ref{table:qg-or} shows the results, which verify that both the sequential states and graph states from the encoding process contribute to the overall performance as removing any one of them causes a performance drop on both F1$_{BLEU1}$ and F1$_{BLEU4}$. Especially, when removing $GS/SS$, those two matrices drops by a great margin, which shows the contributions. The results indicate that bridging the gap between decision making and question generation is necessary.\footnote{Our method is also applicable to other generation architectures such as T5 \cite{raffel2020exploring}. For the reference of interested readers, we tried to employ T5 as our backbone, achieving better performance: 53.7/45.0 for dev and 52.5/43.7 for test (F1BLEU1/F1BLEU4).}

\begin{table}
\small
\centering
\setlength{\tabcolsep}{6.8pt}
\begin{tabular}{lcccc}
\toprule
DPR++ & Top1 & Top5 & Top10 & Top20 \\ 
 \midrule
Train & 84.2 & 99.0 & 99.9  & 100 \\
Dev & 66.3 & 90.0 & 92.4 & 94.5  \\
\quad Seen Only  & 84.6  & 98.0 & 99.8 & 100 \\
\quad Unseen Only & 51.2 & 83.3 & 86.3 & 100 \\
Test & 79.8 & 95.4 & 97.1 & 97.5 \\
\quad Seen Only & 83.7 & 98.5 & 99.9 & 100 \\
\quad Unseen Only & 76.9 & 93.1 & 95 & 95.6 \\
\bottomrule
\end{tabular}
\caption{Retrieval Results of DPR++. }\label{table:or-tfdpr}
\end{table}

\begin{table}
\small
\setlength{\tabcolsep}{1pt}
\centering\centering
\setlength{\tabcolsep}{4.2pt}
\begin{tabular}{lcccc}
\toprule
\multirow{2}{*}{Model} &
\multicolumn{2}{c}{Dev Set} & \multicolumn{2}{c}{Test Set}\\
 & $\text{F1}_{\text{BLEU1}}$ & $\text{F1}_{\text{BLEU4}}$ & $\text{F1}_{\text{BLEU1}}$ & $\text{F1}_{\text{BLEU4}}$ \\ \midrule
 \textsc{Oscar} & 51.3\scriptsize{$\pm$0.8} & 43.1\scriptsize{$\pm$0.8}& 49.1\scriptsize{$\pm$1.1}& 41.9\scriptsize{$\pm$1.8}\\
 \quad w/o GS & 50.9\scriptsize{$\pm$0.9} & 43.0\scriptsize{$\pm$0.7} & 48.7\scriptsize{$\pm$1.3} & 41.6\scriptsize{$\pm$1.5}\\
 \quad w/o SS & 50.6\scriptsize{$\pm$0.6} & 42.8\scriptsize{$\pm$0.5} & 48.1\scriptsize{$\pm$1.4} & 41.4\scriptsize{$\pm$1.4} \\
 \quad w/o both & 49.9\scriptsize{$\pm$0.8} & 42.7\scriptsize{$\pm$0.8} & 47.1\scriptsize{$\pm$1.7} & 40.4\scriptsize{$\pm$1.8}\\
\bottomrule
\end{tabular}
\caption{Question generation results on the OR-ShARC dataset. SS and GS denote the sequential states and graph states, respectively. }\label{table:qg-or}
\end{table}

\paragraph{Smoothing Strategies} We explore the performance of different strategies when fusing the contextual states into BART decoder, and the results are shown in Table \ref{table: strategies}, from which we see that the gating mechanism yields the best performance. The most plausible reason would be the advantage of using the gates to filter the critical information.

\begin{figure*}[!t]
\centering
\includegraphics[width=1\textwidth]{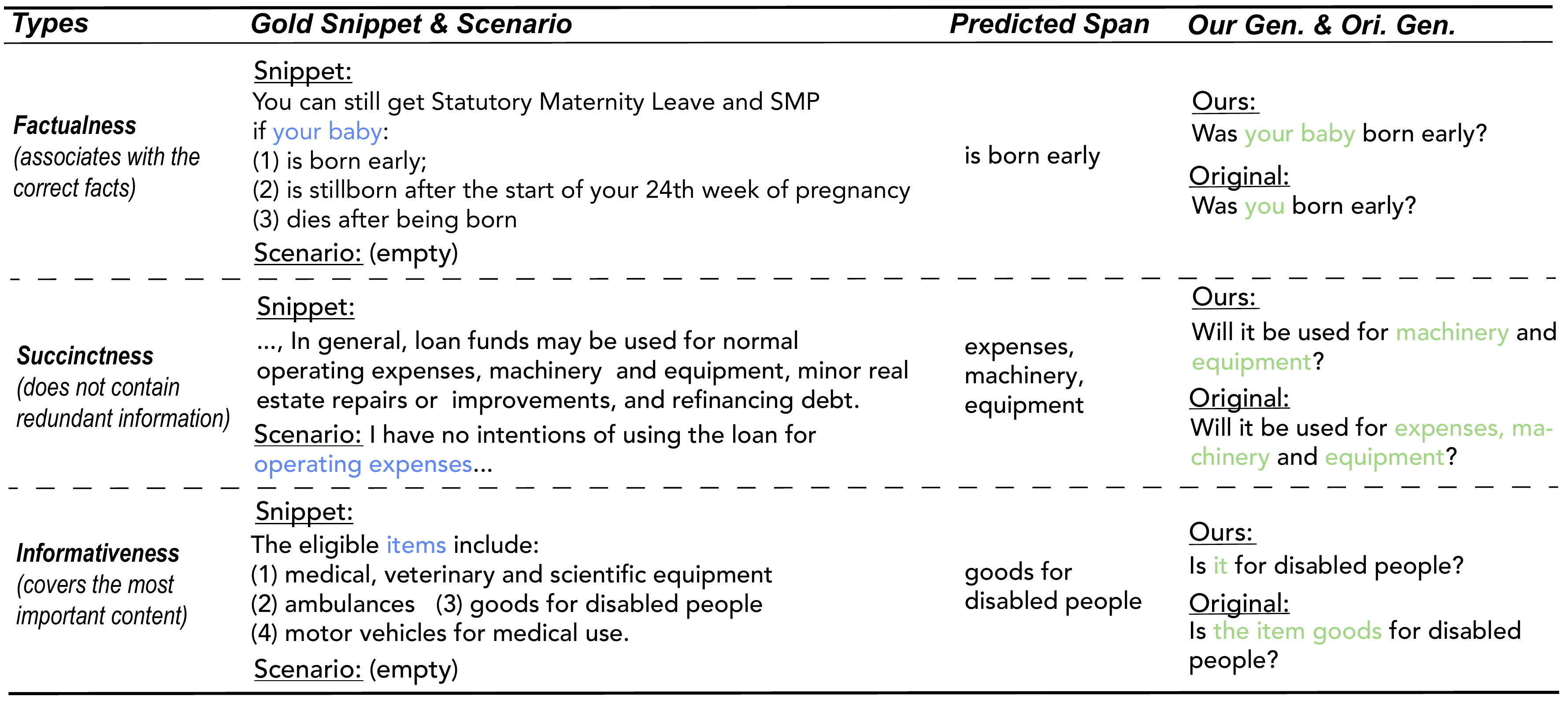}
\caption{Question generation examples of \textsc{Oscar} and the original model.``Our Gen.'' stands for the question generated by \textsc{Oscar}; ``Ori. Gen.'' stands for the question generated by the baseline model.}
\label{qa-examples}
\end{figure*}

\paragraph{Upper-bound Evaluation} To further investigate how the encoder states help generation, we construct a ``gold" dataset as the upper bound evaluation, in which we replace the reference span with the ground-truth span by selecting
the span of the rule text which has the minimum edit distance with the to-be-asked follow-up question, in contrast to the original span that is predicted by our model. We find an interesting observation that the BLEU-1 and BLEU-4 scores drop from $90.64 \rightarrow 89.23$, and  $89.61 \rightarrow 85.81$ after aggregating the DM states on the constructed dataset. Compared with the experiments on the original dataset, the performance gap shows that using embeddings from the decision making stage would well fill the information loss caused by the span prediction stage, and would be beneficial to deal with the errors propagation.

\begin{table}
\small
\setlength{\tabcolsep}{1pt}
\centering\centering
\setlength{\tabcolsep}{4.2pt}
\begin{tabular}{lcccc}
\toprule
\multirow{2}{*}{Model} &
\multicolumn{2}{c}{Dev Set} & \multicolumn{2}{c}{Test Set}\\
 & $\text{F1}_{\text{BLEU1}}$ & $\text{F1}_{\text{BLEU4}}$ & $\text{F1}_{\text{BLEU1}}$ & $\text{F1}_{\text{BLEU4}}$ \\ \midrule
Concatenation & 51.3\scriptsize{$\pm$0.8} & 43.1\scriptsize{$\pm$0.8} & 49.1\scriptsize{$\pm$1.1} & 41.9\scriptsize{$\pm$1.8}\\
Gated Attention & 51.6\scriptsize{$\pm$0.6} & 44.1\scriptsize{$\pm$0.5} & 49.5\scriptsize{$\pm$1.2} & 42.1\scriptsize{$\pm$1.4} \\
\bottomrule
\end{tabular}
\caption{Question generation results using different smoothing strategies on the OR-ShARC dataset.}\label{table: strategies}
\end{table}

\paragraph{Closed-book Evaluation} 
Besides the open-retrieval task, our end-to-end unified modeling method is also applicable to the traditional CMR task. We conduct comparisons on the original ShARC question generation task with provided rule documents to evaluate the performance. Results in Table \ref{table: close-or} show the obvious advantage on the open-retrieval task, indicating the strong ability to extract key information from noisy documents.

\subsection{Case Study}

To explore the generation quality intuitively, we randomly collect and summarize error cases of the baseline and our models for comparison. Results of a few typical examples are presented in Figure. \ref{qa-examples}. We evaluate the examples in term of three aspects, namely, \textit{factualness}, \textit{succinctness} and \textit{informativeness}. The difference of generation by \textsc{Oscar} and the baseline are highlighted in green, while the blue words are the indication of the correct generations. One can easily observe that our generation outperforms the baseline model regarding factualness, succinctness, and informativeness. This might be because that the incorporation of features from the decision making stage can well fill in the gap of information provided for question generation.


\begin{table}
\small
\setlength{\tabcolsep}{1pt}
\centering\centering
\setlength{\tabcolsep}{4.2pt}
\begin{tabular}{lcccc}
\toprule
\multirow{2}{*}{Model} &
\multicolumn{2}{c}{ShARC} & \multicolumn{2}{c}{OR-ShARC}\\
 & \text{BLEU1} & \text{BLEU4} & $\text{F1}_{\text{BLEU1}}$ & $\text{F1}_{\text{BLEU4}}$ \\ \midrule
 \textsc{Baseline} & 62.4\scriptsize$\pm 1.6$  & 47.4\scriptsize$\pm1.6$&50.2\scriptsize$\pm 0.7$ &42.6\scriptsize$\pm 0.5$ \\
\textsc{Oscar} & 63.3\scriptsize$\pm 1.2$  &  48.1\scriptsize$\pm 1.4$& 51.6\scriptsize$\pm 0.6$&44.4\scriptsize$\pm0.4$  \\
\bottomrule
\end{tabular}
\caption{Performance comparison on the dev sets of the closed-book and open-retrieval tasks.}\label{table: close-or}
\end{table}
\section{Conclusion}
In this paper, we study conversational machine reading based on open-retrieval of supporting rule documents, and present a novel end-to-end framework \textsc{Oscar} to enhance the question generation by referring to the rich contextualized dialogue states that involve the interactions between rule conditions, user scenario, initial question and dialogue history. Our \textsc{Oscar} consists of three main modules including retriever, encoder, and decoder as a unified model. Experiments on OR-ShARC show the effectiveness by achieving a new state-of-the-art result. Case studies show that \textsc{Oscar} can generate high-quality questions compared with the previous widely-used pipeline systems.

\section*{Acknowledgments}
We thank Yifan Gao for providing the sources of MUDERN \cite{gao2021open} and valuable suggestions to help improve this work.

\bibliography{custom}
\bibliographystyle{acl_natbib}

\end{document}